\documentclass{article}

\usepackage[accepted]{icml2021}

\usepackage{microtype}
\usepackage{graphicx}
\usepackage{subfigure}
\usepackage{booktabs} 

\icmltitlerunning{Interpretable Machine Learning: Moving From Mythos to Diagnostics}

\begin{document}

\twocolumn[
\icmltitle{Interpretable Machine Learning: Moving From Mythos to Diagnostics}



\icmlsetsymbol{equal}{*}
\icmlsetsymbol{equal2}{$\dagger$}

\begin{icmlauthorlist}
\icmlauthor{Valerie Chen}{equal,cmu}
\icmlauthor{Jeffrey Li}{equal,uw}
\icmlauthor{Joon Sik Kim}{equal2,cmu}
\icmlauthor{Gregory Plumb}{equal2,cmu}
\icmlauthor{Ameet Talwalkar}{cmu,dai}
\end{icmlauthorlist}

\icmlaffiliation{cmu}{Carnegie Mellon University}
\icmlaffiliation{dai}{Determined AI}
\icmlaffiliation{uw}{University of Washington}

\icmlcorrespondingauthor{Valerie Chen}{valeriechen@cmu.edu}
\icmlcorrespondingauthor{Jeffrey Li}{jwl2162@cs.washington.edu}

\icmlkeywords{Machine Learning, ICML}

\vskip 0.3in
]



\printAffiliationsAndNotice{\icmlEqualContribution} 

\begin{abstract}
Despite years of progress in the field of Interpretable Machine Learning (IML), a significant gap persists between the technical objectives targeted by \emph{researchers' methods} and the high-level goals stated as \emph{consumers' use cases}. To address this gap, we argue for the IML community to embrace a \emph{diagnostic} vision for the field. Instead of viewing IML methods as a panacea 
for a variety of overly broad use cases, we emphasize the need to 
systematically connect IML methods to narrower--yet better defined--target use cases. To formalize this vision, we propose a taxonomy including both methods and use cases, helping to conceptualize the current gaps between the two. 
Then, to connect these two sides, we describe a three-step workflow to enable researchers and consumers to define and validate IML methods as useful diagnostics. Eventually, by applying this workflow, a more complete version of the taxonomy will allow consumers to find relevant methods for their target use cases and researchers to identify motivating use cases for their 
methods.
\end{abstract}

\section{Introduction}

The emergence of machine learning as a society-changing technology in the last decade has triggered concerns about our inability to understand the reasoning of increasingly complex models. The field of Interpretable Machine Learning (IML)\footnote{The literature sometimes differentiates \textit{interpretable} ML (i.e., designing models which are understandable by-design) and \textit{explainable} ML (i.e., producing post-hoc explanations for models) \cite{rudin2019stop}.
We emphasize that whether an explanation is produced by-design or by a post-hoc method does not affect how it should be used or evaluated (though it may affect the quality of the results). Thus, we see this distinction as orthogonal to our paper.} 
grew out of these concerns, with the goal of empowering various stakeholders to tackle use cases such as building trust in models, performing model debugging, and generally informing real human-decision making \cite{bhattpaper,lipton2018mythos, gilpin2018explaining}.



However, despite the flurry of IML methodological development over the last several years, a stark disconnect characterizes the current overall approach: IML methods typically optimize diverse but narrow technical objectives, yet their claimed use-cases remain broad and often under-specified. 
Echoing similar critiques about the field made \cite{lipton2018mythos}, it has thus remained difficult for the field to sufficiently evaluate these claims 
and thus to translate 
methodological advances into widespread practical impact.

In this paper, we outline a path forward for the ML community to address this disconnect and foster more widespread adoption, focusing on two key principles:

\begin{figure}[]
\centering
  \includegraphics[scale=0.25]{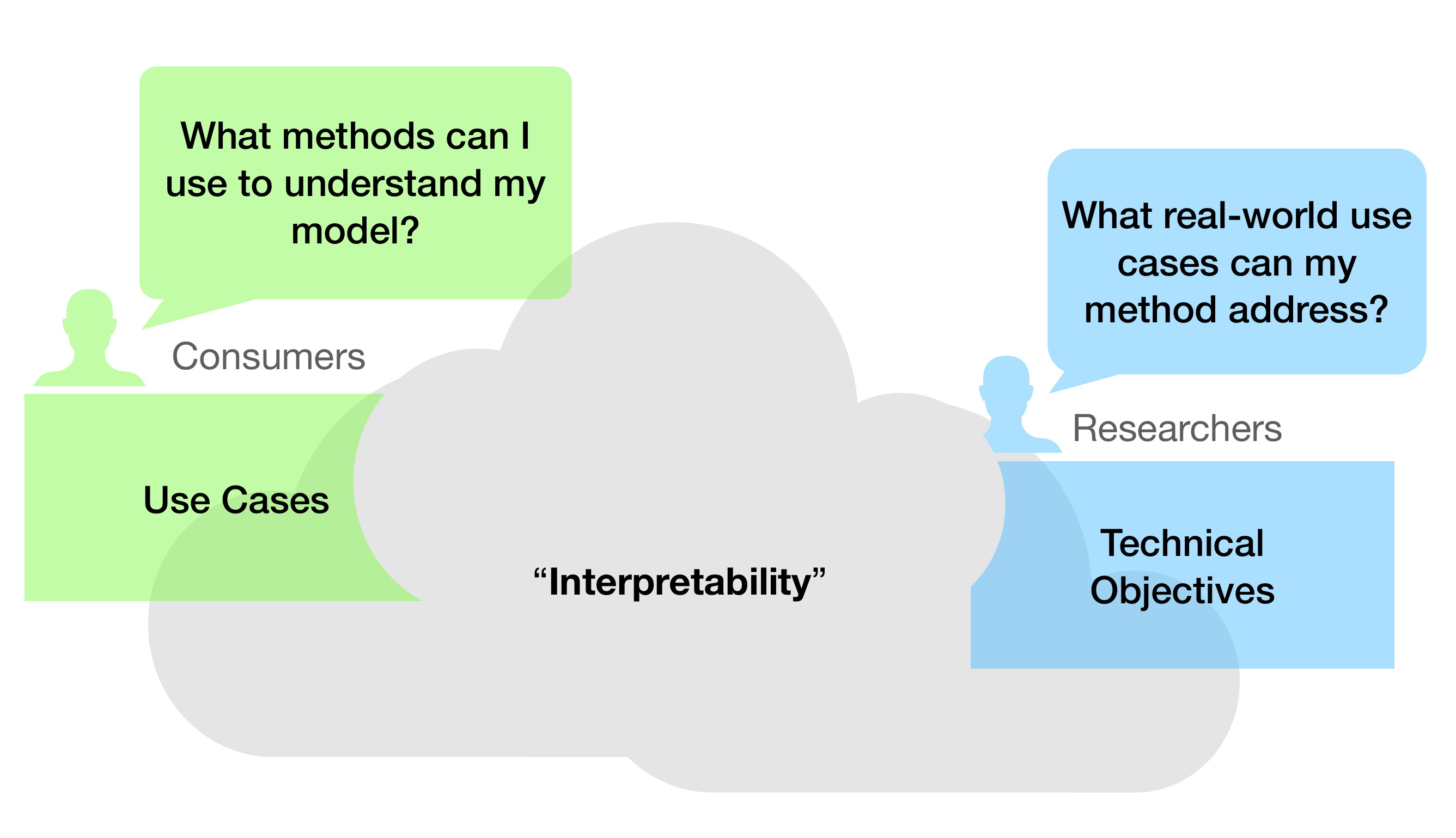}
  \caption{Currently, IML researchers focus more on technical objectives while consumers focus on use cases. Often, a lack of explicit connections remains between the two, making proper usage and development of IML methods difficult for both parties. 
  }
  \label{fig:teaser}
  \vspace{-0.75cm}
\end{figure}

\begin{figure*}[]
\centering
  \includegraphics[scale=0.66]{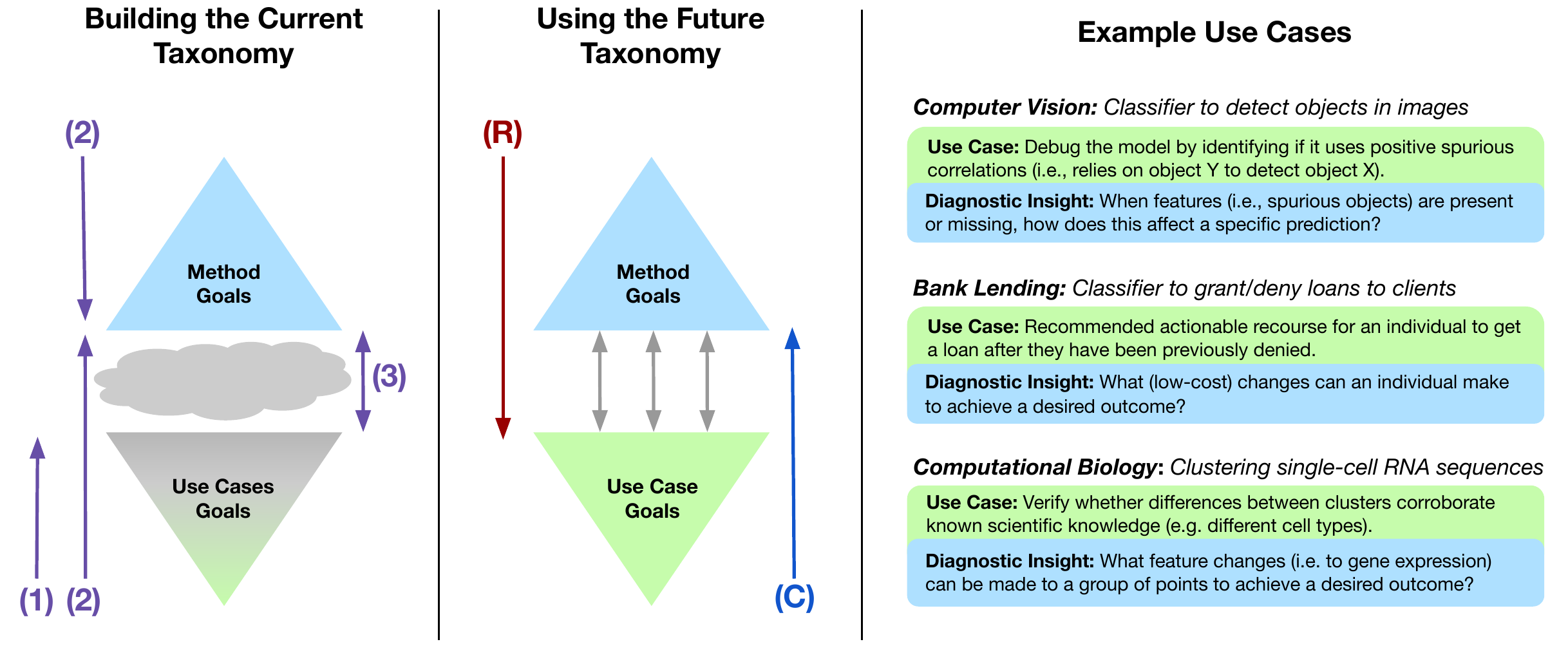}
  \caption{(Left) We focus on how researchers and consumers can work together to both establish a better use case organization (i.e., ``Use Case Goals'') and further connections through the current gap between methods and use cases (i.e., the cloud) by following steps (1)-(3) in our proposed workflow. 
  (Middle) As the two sides are increasingly connected to one another, researchers (R) and consumers (C) can make use of the taxonomy to find use cases for their methods and methods for their use cases, respectively. (Right) We highlight how three different potential diagnostics may provide useful insights for three use cases. In fact, the desired diagnostic information in each use case maps to a different Technical Objective (local feature attribution, local counterfactual, and global counterfactual, respectively) in our taxonomy (Figure \ref{fig:taxonomy}). When we later discuss a more concrete workflow for filling in the taxonomy we expand on the computer vision setting as a running example.}
  \label{fig:abstractedversion}
\end{figure*}


\textbf{1. Embrace a “diagnostic” vision for IML.} 
Instead of aiming to provide complete solutions for ill-defined problems such as ``debugging'' and ``trust'', we argue that the field of IML should focus on the important, if less grandiose, goal of developing a suite of rigorously-tested diagnostic tools. In treating IML methods as future diagnostics, we view each as providing a targeted, well-specified insight into a model’s behavior. In this sense, these methods should then be used alongside and in a manner similar to more classical statistical diagnostics (e.g., error bars, hypothesis tests, methods for outlier detection), for which clearer guidelines exist for when and how to apply them.\footnote{Under this vision, we treat existing IML methods as \textit{potential} diagnostics, emphasizing their need to be more rigorously-tested.}

\textbf{2. Rigorously evaluate and establish potential IML diagnostics.}
Currently, IML researchers typically develop and evaluate methods by focusing on quantifiable technical objectives, e.g., maximizing various notions of faithfulness or adherence to some desirable axioms \cite{lundberg2017unified, sundararajan2017axiomatic, bach2015pixel}. However, while these IML methods generally target seemingly relevant aspects of a model’s behavior, it is imperative to carefully measure their effectiveness on concrete use cases in order to demonstrate their utility as practical diagnostics.



Motivated by these principles, we first illustrate our diagnostic vision via an incomplete taxonomy that synthesizes 
foundational works on IML methods and evaluation.
The taxonomy (as shown at an abstract level in Figure \ref{fig:abstractedversion} and discussed in more depth in Section \ref{section:taxonomy}) not only serves as a template for building an explicit mapping between potential IML diagnostics and specific use cases, but also as a tool to unify studies of IML's usefulness in real-world settings (concrete examples shown in Figure \ref{fig:abstractedversion}, right). 

However, the incompleteness of  
the current taxonomy emphasizes the need for researchers and consumers 
to work together to expand its coverage and refine connections within it. 
More specifically, doing so requires careful considerations at each of these 3 steps of the IML workflow in the context of our taxonomy as shown in Figure \ref{fig:abstractedversion} (left):
\newpage 
\begin{enumerate}
    \item[(1)] \emph{Problem Definition}, where researchers work with consumers to define a well-specified \textit{target use case} (TUC).
    \item[(2)] \emph{Method Selection}, where they identify potential IML methods for a TUC by navigating the methods part of the taxonomy \textit{and/or} leveraging previously established connections between similar use cases and methods.
    \item[(3)] \emph{Method Evaluation}, where they test whether selected methods can meet TUCs. 
\end{enumerate}
\vspace{-0.2cm}
In Section \ref{section:workflow}, 
we provide an extensive discussion about best practices for this IML workflow to flesh out this taxonomy and deliver rigorously-tested diagnostics to consumers. 
Ultimately, we envision an increasingly complete taxonomy that
(i) allows consumers to find suitable IML methods for their use cases; and (ii) helps researchers to ground their technical work in real applications (Figure \ref{fig:abstractedversion}, middle). 



\section{Background} \label{sec:relatedwork}
An increasingly diverse set of methods has been recently proposed and broadly classified as part of IML. 
However, multiple concerns have been expressed in light of this rapid development, focused on IML's underlying foundations and the gap between research and practice. 

\textbf{Critiques of the field’s foundations:} 
\cite{lipton2018mythos} provided an early critique, highlighting that the stated motivations of IML were both highly variable,
and potentially discordant with proposed methods. 
\cite{krishnan2019against} added to these arguments from a philosophical angle, positing that interpretability as a unifying concept is both unclear and of questionable usefulness. Instead, as they argue, more focus should be placed on the actual end-goals, for which IML is one possible solution. 

 
\textbf{Gaps between research and practice:} Multiple works have also highlighted important gaps between existing methods and their practical usefulness. Some have demonstrated a lack of stability/robustness of popular approaches \cite{adebayo2018sanity, laugel2019issues, alvarez2018robustness}. Others discuss how common IML methods can fail to help humans in the real-world, both through pointing out hidden assumptions and dangers \cite{barocas2020hidden, rudin2019stop} as well as conducting case-studies with users \cite{bansal2020does, kaur2020interpreting}.
 
More recently, many review papers \cite{gilpin2018explaining, mohseni2019multidisciplinary, murdoch2019interpretable, arya2019one} have attempted to clean up and organize aspects of IML, but largely do not address these issues head-on. 
In contrast, our proposed re-framing of IML methods as diagnostic tools follows naturally from these concerns.
Notably we embrace the seeming shortcomings of IML methods as providing merely “facts” \cite{krishnan2019against} or “summary statistics” \cite{rudin2019stop} about a model, and instead focus on the practical questions of when and how these methods can be practically useful.



\section{A Diagnostic Vision for IML}

\label{section:taxonomy}

\begin{figure*}[!t]
    \centering
    \includegraphics[width=1.0\textwidth]{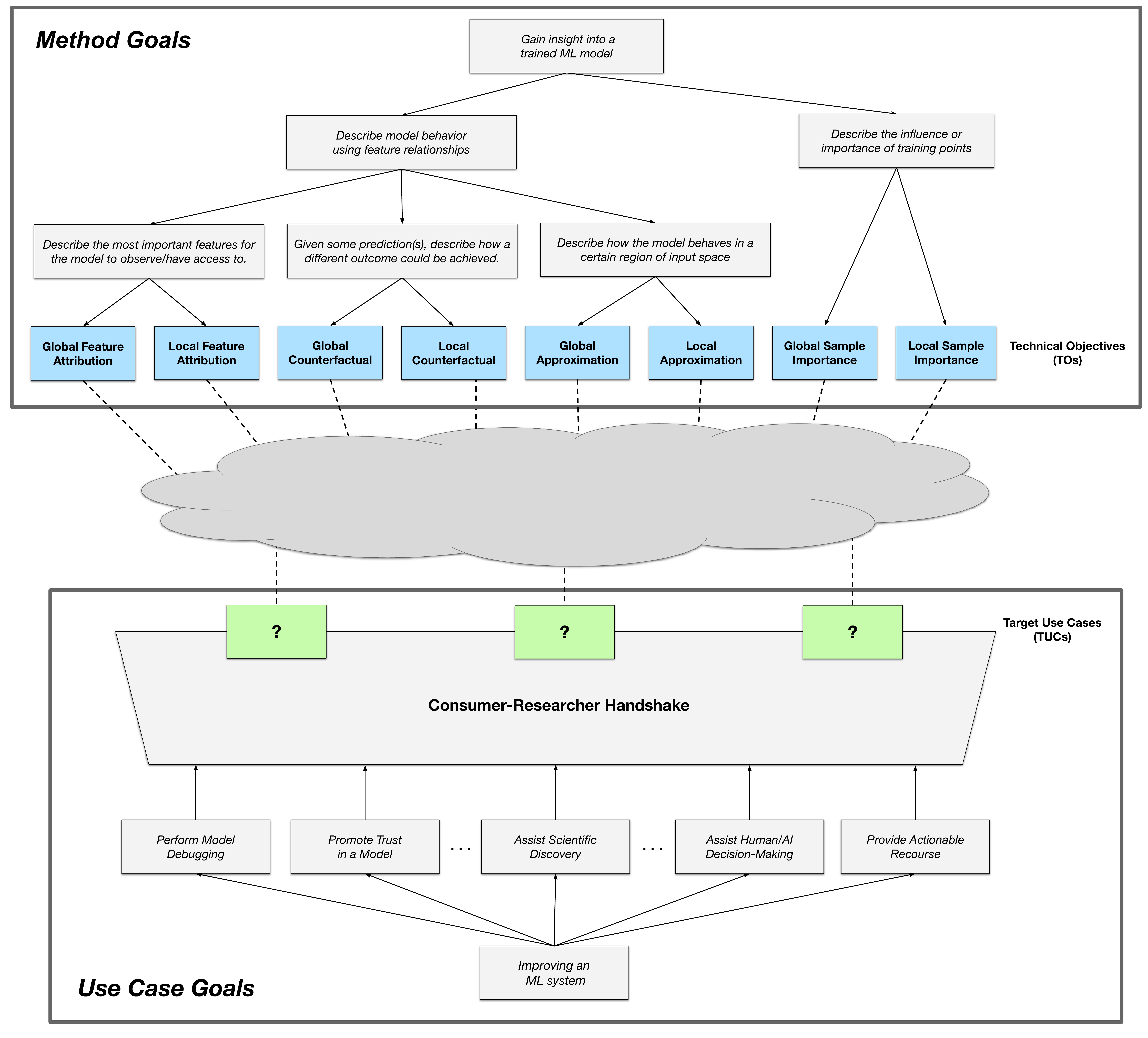}
    \caption{Our taxonomy consists of a hierarchical organization of both existing method and use case goals. Moving forwards, the goal for researchers and consumers is to conduct principled studies to both refine the current organization of use cases by defining more well-specified target use cases (green) and to establish explicit connections between these targets and technical objectives (blue). 
    }
    \label{fig:taxonomy}
\end{figure*}

We think of a diagnostic as a tool that provides some insight about a model. As an analogy, consider the suite of diagnostic tools at a doctor's disposal
that similarly provides insight about a patient. An x-ray could be useful for identifying bone fractures while a heart rate monitor would be helpful for identifying an irregular heart rhythm. Importantly, neither tool enables the doctor to broadly ``understand'' a person's health.  However, each can be useful \textit{if applied properly to a well-scoped problem}. Similarly, rigorously establishing connections between IML methods and well-defined use cases is imperative for the IML community.

To begin such a pursuit, we start by first identifying and reconciling the many method goals and use case goals that one might encounter currently. Based on contemporary practices and discourse, we propose a taxonomy that organizes separately the method goals at the top-end and use case goals at the bottom-end (Figure~\ref{fig:taxonomy}). While our diagnostic vision for the field ideally involves a robust set of connections between these two sides, we use a ``cloud'' to illustrate the current overall lack of well-established diagnostics. 




\subsection{Method Goals} 
\label{sec:methodgoals}

Each IML method provides a specific type of insight into a given model. Based on these types of insights, we first provide a hierarchical organization which divides the set of existing IML methods into 8 method clusters. In the diagnostic vision, we think of each method cluster broadly as a class of diagnostics that addresses a Technical Objective (TO). Then, we describe in more detail each TO in a way that allows one to specify individual method objectives.

\subsubsection{Hierarchical Organization} 
The top-end of our taxonomy aims to differentiate between the various perspectives explanations provide based on three factors commonly discussed in existing literature \cite{arya2019one, guidotti2018survey, doshivelez2017rigorous}. We discuss these further in Appendix \ref{appendix:taxonomy}.

At the leaf nodes are \textit{technical objectives} (TOs), classes of goals that are precise enough to be generally linked to a \textit{method cluster} that most directly addresses them.
In total, there are 8 TOs/method clusters which captures a large portion of the goals of current IML methods. We note a few important nuances regarding our characterization of TOs. 

First, although TOs and method cluster are bijective in our proposed taxonomy, it is important to explicitly distinguish these two concepts because of the potential for \emph{cross-cluster adaptation}.
This notion arises because
it is frequently possible for that method to, in an ad-hoc fashion, be adapted to address a different TO. 

Second, we emphasize that each TO should be thought of as defining a \textit{class} of related goals. 
Indeed, for a given TO, we hypothesize some of the key \textit{technical detail(s)} that must be considered towards fully parametrizing meaningfully different instantiations of the same broader goal. 
These important technical details, taken together with the TO, allow one to define individual \textit{proxy metrics} that reflect the desired properties of one's explanations. Proxy metrics can then serve as tractable objective functions for individual methods to optimize, as well as measures of how well any method addresses a particular instantiation of the TO. 

\subsubsection{Technical Objectives} \label{section:methodclusters} 
We next overview the TOs (and their technical details) that correspond to various method clusters.
Due to the overlaps in content, we group together local and global versions of the same general method type/objective (for more extensive details and examples of specific methods for each, see Appendix \ref{appendix:methodclusters}). 

\textit{\underline{Feature attribution}} address when features are present (or missing), how does this affect the the model's prediction(s) (i.e. how ``important'' each feature is to the model's prediction(s)). Often, measures of importance are defined based on how the model's prediction(s) change relative to its prediction for some baseline input. The baseline input is sometimes implicit and domain specific (e.g., all black pixels for grayscale images or the mean input in tabular data).
Thus, the technical details are both the precise \textit{notion of “importance”} and the choice of the \textit{baseline input}. Relevant proxy metrics typically measure how much the model prediction changes for different types of perturbations applied to the individual (or the training data) according to the ``importance'' values as computed by each method.

\underline{\textit{Counterfactual}} explanations address what “low cost” modification can be applied to data point(s) to achieve a desired prediction. 
The most common technical detail is the specific measure of \textit{cost} and the most common proxy metric is how often the counterfactual changes the model's prediction(s). 



\underline{\textit{Approximation}} methods address how can one summarize the model by approximating its predictions in a region, either locally around a data point, globally around as many points as possible, or across a specific region of the input space. 
These methods require the technical detail of both what that \textit{region} is and what the simple function’s \textit{model family} is. For local approximation, a canonical metric is local fidelity, which measures how well the method predicts within a certain neighborhood of a data point. For global approximation, a proxy metric is coverage, which measures how many data points the explanation applies to.



\underline{\textit{Sample importance}} methods address what training points are influential on a model's prediction for either an individual point or the model as a whole. 
Technical details differ from method to method, so currently it is difficult to identify a uniform axis of variation. 
These methods can be evaluated with proxy metrics that represent the usefulness of the provided explanations, through simulated experiments of finding corrupted data points, detecting points responsible data distribution shifts, and recovering high accuracy with the samples considered important.



\subsubsection{\textbf{How do by-design methods fit in?}} 
While they do not have a corresponding method cluster in our taxonomy, it is important to discuss another family of IML methods called ``interpretable by-design” methods \cite{rudin2019stop}. 
The differentiating property of these models from the post-hoc methods that we reference above is that the TO(s) of these approaches is intrinsically tied to the model family itself, hence the models are interpretable by design. 
That said, by-design methods also fit into our framework and should be viewed as a different way to answer the same TOs in our taxonomy. 
When by-design methods are proposed or used, they should clearly specify which TO(s) they are intending to address.


\subsection{Use Case Goals} 
\label{sec:usecasegoals}

Currently, much of the discourse on IML use cases surrounds differentiating fairly broad goals, such as model debugging, gaining trust of various stakeholders, providing actionable recourse, assisting in scientific/causal discovery, and aiding Human/AI teams \cite{bhattpaper,lipton2018mythos, gilpin2018explaining} (Figure~\ref{fig:taxonomy}).
While this represents a good start, it is of limited utility to treat each of these categories as monolithic problems for IML to solve. For one, these problems are complex and should not be assumed to be completely solvable by IML itself. Rather, IML is but one potential set of tools that must be demonstrated to be useful. That is, to show that an IML method is an effective diagnostic, specific use cases must be identified and demonstrated \cite{krishnan2019against}. 
Secondly, each broad goal really includes multiple separate technical problems, crossed with many possible practical settings and constraints. It is likely that a given IML method will not be equally useful across the board for all of these sub-problems and domains.

Thus, claims of practical usefulness should ideally be specified down to the level of an adequately defined \textit{target use case} (TUC).
TUCs, like TOs on the methods side, correspond to learning a specific relevant characteristic about the underlying model (e.g. a certain property or notion of model behavior). 
However, unlike a TO, they represent real-world problems that, while evaluable, often might not be amenable to direct optimization. For example, one can set up real or simulated evaluations (see Section \ref{section:pitfall3}) to determine whether an IML method is useful for identifying a particular kind of bug in the model (e.g. spurious positive correlations), but it is not so obvious how to optimize an IML method that will succeed on those real or simulated evaluations. 
\section{A Workflow for Establishing Diagnostics}
\label{section:workflow}

We now turn to how a diagnostic vision for IML can be more fully realized, discussing how methods can be established as diagnostics, thus filling gaps in the existing taxonomy. 
Specifically, we define an ideal workflow for consumer-researcher teams to conduct future studies about IML methods describing how the taxonomy can guide best practices for each of the three key steps: (1) Problem Definition, (2) Method Selection, and (3) Method Evaluation. This workflow applies to both teams who wish to study existing IML methods and those who are proposing new ones. 


To help contextualize this discussion, we provide a running example that builds on the Computer Vision model debugging example from Figure~\ref{fig:abstractedversion} (right). Model debugging is not only a common \cite{hong2020human, bhattpaper}, but also well-grounded consumer use case. It is a natural starting point due to the versatile nature of its assumed consumer, data scientists, typically has both substantial ML knowledge and domain expertise, minimizing the communication gap between the data scientist and the IML researcher.


\begin{figure}[]
\centering
  \includegraphics[scale=0.26]{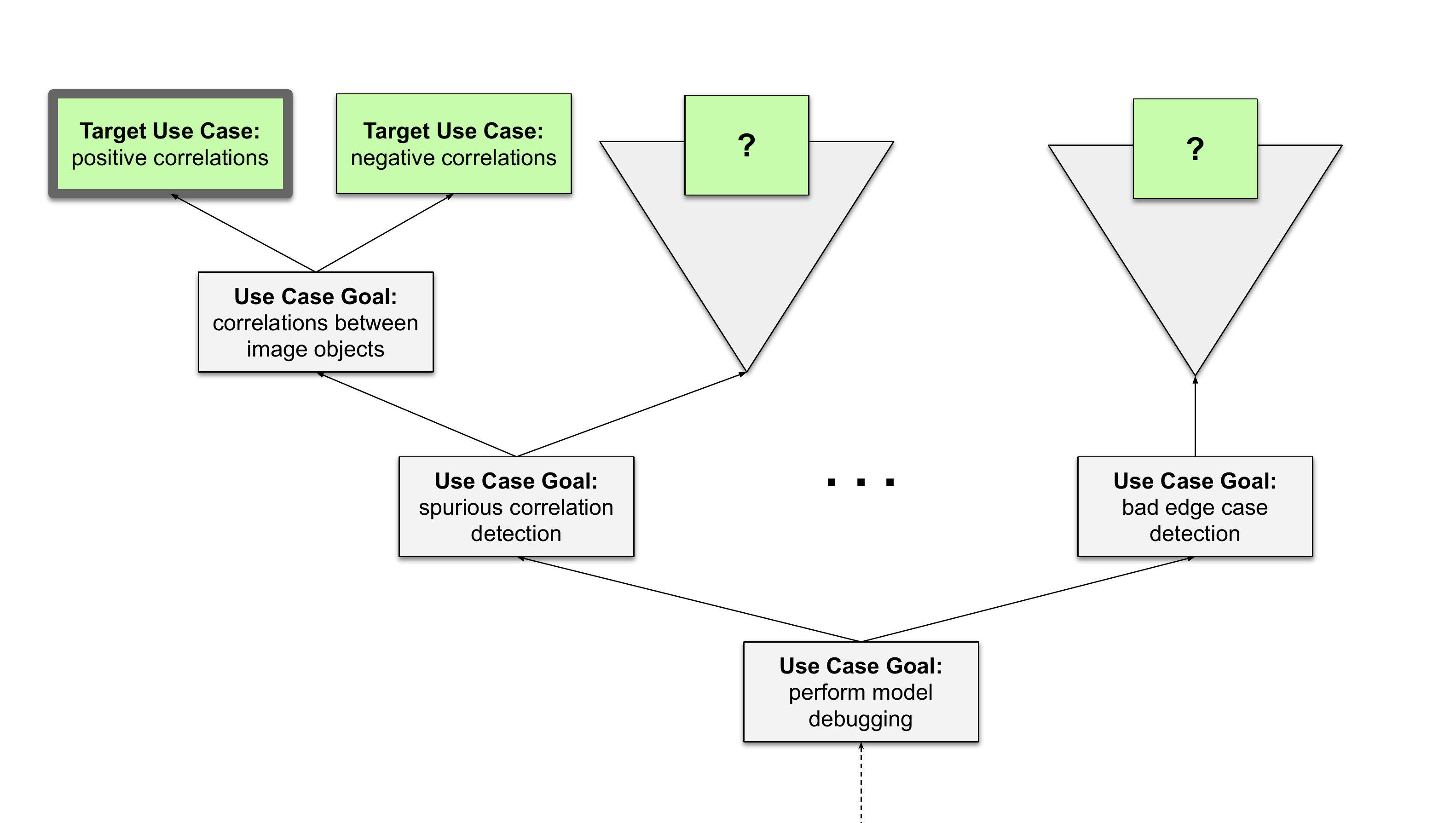}
  \caption{A hypothetical version of the use cases part of our taxonomy as produced by the consumer-researcher handshake in our running example. 
  The identified TUC is highlighted by the box with the thicker border.}
  \label{fig:runningexample}
\end{figure}

\subsection{Step 1: Problem Definition}
\label{section:pitfall1}

As motivated by Section \ref{sec:usecasegoals}, we argue that an important first step for any principled study is to define a well-specified TUC. We call this the \textit{consumer-researcher handshake} (Figure \ref{fig:taxonomy}), where researchers work with consumers to progressively refine the latter's real-world problems into relevant TUCs. In this process, some helpful pieces of information that should be discussed include: the data available, the ML pipeline used, the domain knowledge required to perform evaluations, etc. Ultimately, a more flushed out taxonomy will help researchers to have more concrete use cases at hand to motivate their method development and consumers to have more realistic guidance on what IML can and cannot do for them.

\textit{\textbf{Running Example:} Consider a data scientist who wants to debug their image-based object detection model. 
The team needs to identify a TUC that is more specific than ``perform model debugging'' by identifying exactly what the notion of ``bug'' is that the IML method should detect.
As shown in a hypothetical version of the use cases part of the taxonomy (Figure \ref{fig:runningexample}), the umbrella of model debugging includes sub-problems such as detecting spurious correlations and  identifying bad edge-case behavior.
Through the consumer-researcher handshake, it arises that the data scientist is concerned the model might not be making correct decisions based on the actual target objects, but rather relying on correlated objects which also happen to be present. 
For example, the model might be using the presence of a person as an indicator that there is a tennis racket in the image, instead of the racket itself.
}

\textit{This information allows the team to navigate the portion of the taxonomy in Figure \ref{fig:runningexample}.
By considering the data scientist's concern, they first narrow the goal from model debugging to detecting spurious correlations (SCs). 
Then, by also taking into account the specific setting (i.e. the presence of the tennis racket at the same time as the tennis player), they are able to arrive at a further specified use case of detecting SCs between two positively correlated objects. 
In this case, the team takes care to differentiate this from the analogous problem of detecting reliance on negatively correlated objects, reasoning that the latter is fundamentally different (i.e., it is harder to tell that the output depends on an object or not if the co-occurrences are rare in the first place).}

\subsection{Step 2: Method Selection}
\label{section:pitfall2}


After a TUC has been properly defined, the next step is to consider which IML methods might be appropriate. 
This does assume that IML methods are necessary, that is the team should have demonstrated that the TUC presents challenges to more ``trivial'' or conventional diagnostics. For example, \cite{bansal2020does} found model confidence to be a competitive baseline against dedicated interpretability approaches for AI-human decision making teams.

If non-IML diagnostics are unsuccessful, there are two ways the taxonomy can be used to select methods.
First, researchers and consumers can, as a default, traverse the methods part of the taxonomy to hypothesize the TOs (and thus respective method clusters) that might best align with the TUC.
Doing so should rely on the researcher's best judgment in applying prior knowledge and intuition about various method types to try to narrow down the set of potential TOs. If a method is being proposed, the method should be mapped to the appropriate method cluster and the same selection process should follow.
Second, the team can also navigate starting from the use cases part, leveraging and expanding on connections established by previous studies. 
Naturally, if some methods have already been shown to work well on a TUC, then those (or similar) methods provide immediate baselines when studying the same (or similar) use cases.

In either case, an important --yet subtle-- choice must then be made for each method: exactly how its resulting explanations should be interpreted, i.e. which TO is being addressed. 
As discussed in Section \ref{sec:methodgoals}, a method belonging to a specific cluster may most naturally address the associated TO, but it is also possible, and indeed commonplace, to attempt \textit{cross-cluster adaptation} for addressing other TOs. 
Unfortunately, while such adaptations are perhaps useful at times, they are often performed in an ad hoc fashion. 
Specifically, the differences between the technical details of each TO are often overlooked in the adaptation process, which we illustrate next via two examples (and in more depth in Appendix \ref{appendix:crosscluster}). 

\begin{figure*}[!t]
    \centering
    \includegraphics[width=1.0\textwidth]{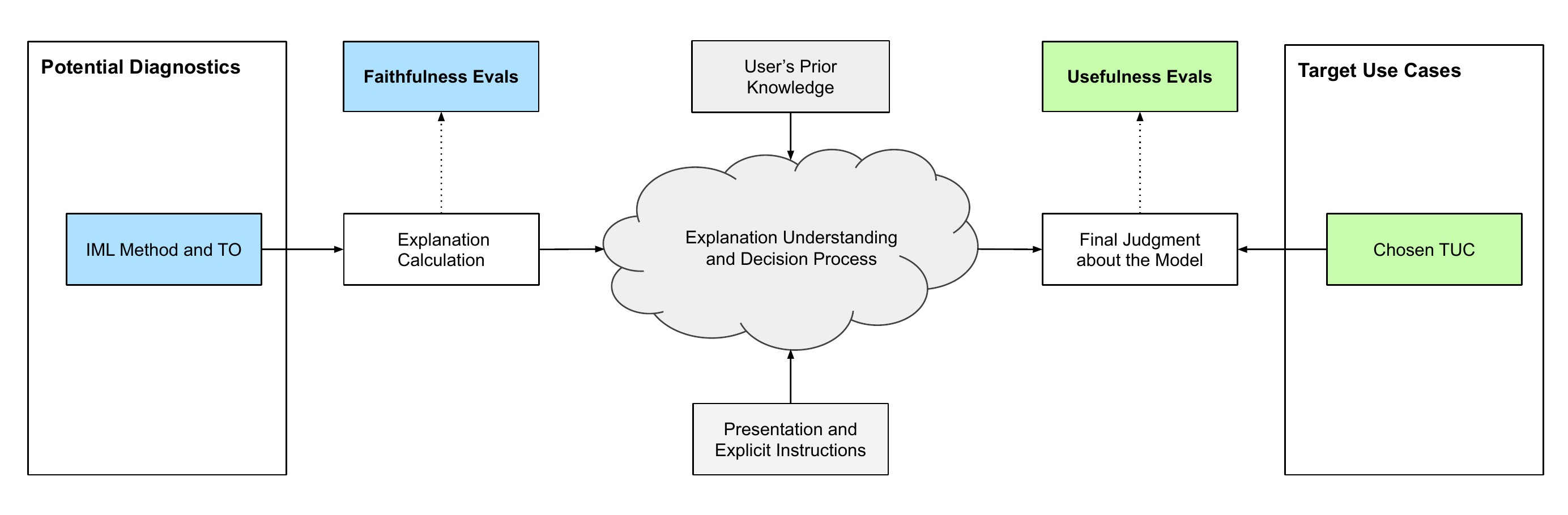}
    \caption{An overview of where different types of IML evaluations (faithfulness and usefulness) fall within the overall pipeline of IML applications. We highlight differences in goals of these evaluations as well as the various moving components that affect them. Colored boxes denote components that need to be defined (IML method and TO, Chosen TUC), while gray boxes to denote components that require more careful study (Explanation Understanding and Decision Process). 
    }
    \label{fig:evaluations}
\end{figure*}


First, one might try to use feature importance weights, via SHAP \cite{lundberg2017unified}, as linear coefficients in a local approximation. 
Such an adaptation assumes that the notion of local ``importance'' also can reflect linear interactions with features on the desired approximation region. However, this is not necessarily guaranteed by SHAP, which instead enforces a different set of game-theoretic desiderata on the importance values and may be set up to consider a quite disparate set of perturbations compared to the target approximation region.  

Conversely, one can think of saliency maps via vanilla gradients~\cite{simonyan2013deep} as an adaptation in the opposite direction. 
These saliency maps, a local approximation where the effective neighborhood region is extremely small, are more popularly used to address local feature attribution objectives such as to identify which parts of the image are affecting the prediction the most.
However, this adaptation carries an underlying assumption that the pixels with the largest gradients are also the most ``important''. This approximation may not be accurate because the local shape measured by the gradient is not necessarily indicative of the model's behavior near a baseline input that is farther away. 


\textit{\textbf{Running example:} In this scenario, suppose that there have not been previously established results for detecting positive SCs. 
The team follows the methods part of the taxonomy to generate hypotheses for which types of local explanations best suits their needs for understanding individual images. 
They decide against approximation based objectives, because as the inputs vary in pixel space, simple approximations are unlikely to hold or be semantically meaningful across continuous local neighborhoods.
They choose feature attribution because they hypothesize that visualizing the features that the model deems most important would be useful for detecting these types of SCs.} 

\textit{The team proposes a method in the local counterfactual method cluster that identifies the super-pixels that must change in order to flip the prediction from ``tennis racket'' to ``no tennis racket''. 
By ``visualizing'' the counterfactual explanation like a saliency map, the team performs a cross-cluster adaptation to interpret the counterfactual as a feature attribution explanation. 
To do so, they are assuming that the most changed features are also the most important to detecting the tennis racket. 
They reason that a feature attribution explanation would be a more intuitive format for the data scientist for this TUC. 
In terms of comparison, a feature attribution method that the team selects for comparison is Grad-CAM \cite{selvaraju2017grad}, which also produces a saliency map.}




\subsection{Step 3: Method Evaluation}
\label{section:pitfall3}

Once appropriate method(s) have been chosen, the last step is to evaluate them. Evaluation is the crucial step of testing whether proposed methods can actually help address the specified TUC. However, evaluations are often carried out in manners incongruent with the properties they claim to test. One common mistake is that the evaluation of an explanation’s \textit{faithfulness} (i.e. ability to meet a specified TO) is often problematically conflated with the evaluation of its \textit{usefulness} (i.e., applicability for addressing practical TUCs). While both may play important roles, as we discuss further in this section, they target fundamentally different claims. 



Our taxonomy addresses this mistake by mapping these evaluations to its different components: faithfulness corresponds to meeting objectives of a specific TO in the methods part and usefulness corresponds to meeting the TUC in the use cases part. Next, using Figure \ref{fig:evaluations} as a guide, we clarify differences between these two types of evaluations and how they can be carried out. 

\textbf{Faithfulness Evaluations} are performed 
with respect to a proxy metric specified using the relevant technical details from the target TO class. For example, if the goal was to show the usefulness of an approximation-based explanation adapted as a counterfactual, the faithfulness evaluation should be with respect to a counterfactual proxy metric. Referring to the terminology from \cite{doshivelez2017rigorous}, these types of evaluations are called \textit{functionally-grounded}, that is involving automated proxy tasks and no humans. While such evaluations are easiest to carry out, they come with key limitations.

In general, one should expect that a method would perform well at least on a proxy for its selected TO and, naturally, those methods which do not directly target this specific proxy will likely not perform as well. An explanation can also be faultily compared as a result of unfair or biased settings of technical details. 
As an example, although GAMs \cite{hastie1990generalized} and linear models both provide local approximations, comparing these methods only in the context of fidelity ignores the fact that GAMs potentially generate more ``complicated'' explanations. 

Further, while faithfulness evaluations can act as a first-step sanity check before running more costly usefulness evaluations, showing that a method is faithful to the model alone is not conclusive of the method's \textit{real-world} usefulness until a direct link is established between the corresponding proxy and TUC. 
Once these links are established, these proxies can then be used more confidently to help rule out bad set-ups before performing expensive usefulness evaluations.

\textbf{Usefulness evaluations}, in contrast to faithfulness, measure a user’s success in applying explanations on the specified TUC. Since they are ultimately an evaluation of what one \textit{does} with an explanation, usefulness depends crucially on factors such as users' prior knowledge, such as domain and ML/IML experience. Referring to the terminology from \cite{doshivelez2017rigorous}, users' perspectives can be incorporated through studies on real humans performing simplified or actual tasks (i.e. \textit{human-grounded} or \textit{application-grounded} evaluations respectively). In particular, to successfully utilize explanations in practice, we would need to study how this process might differ depending on the presentation of the explanation and explicit instructions that are provided. 

As highlighted by the cloud in Figure \ref{fig:evaluations}, how exactly users translate explanation calculations (in their minds) to their final judgments remains murky. This motivates further research relating to better understanding \textit{what users understand explanations to tell them} and \textit{how they act upon these understandings}. Then, when establishing new diagnostics, these assumptions/limitations should be clearly spelled out for when researchers use the method in a future study and when the consumers deploy the method.

Motivated by these challenges, we suggest researchers consider another type of usefulness evaluation called \textit{simulation evaluation}. Simulation evaluation is an algorithmic evaluation on a simulated version of the real task where success and failure is distilled by a domain expert into a measurable quantity (as illustrated in the running example).
This type of evaluation is still based on the real task, but is easier and potentially more reliable to run than user studies. By simulating the users and their decision-making process algorithmically, thus controlling some noisier aspects of usefulness evaluation, researchers may be able to better understand why their methods are “failing”. Is it because of the algorithm itself, or the actual decision process users take? 

Overall, success on these various levels of evaluations provides evidence for establishing a connection between the method in question and the TUC. Specifically, the team should check to see if the proxy metrics considered earlier were correlated to success on the TUC. If so, this would provide evidence for whether the proxy metrics considered should be used again in future studies, connecting faithfulness and usefulness evaluations.



\textit{\textbf{Running example: } The team first performs respective local feature attribution faithfulness evaluations for both methods using the notions of importance that each defines. For example, for the proposed method, the team ensures that each generated explanation \textit{faithfully} carries out its intended TO of identifying the effect of the presence or missingness of a super-pixel. However, good performance on any proxy metric does not conclusively imply good performance on the \textit{actual} TUC, so they turn to usefulness evaluation.}

\textit{The team first conducts a simulation evaluation, where a set of datasets is created that contains either an (artificially induced) positive correlation between a pair of objects or no such correlations.  
By carefully controlling the training and validation distributions, they can automatically verify whether or not a model has learned the problematic behavior they want to detect. Then, they can define a scoring function for the explanations (i.e., how much attention they pay to the spurious object) and measure how well that score correlates with the ground truth for each explanation.
}

\textit{Second, the team runs a human study with multiple models where they know the ground truth of which ones use SCs. They score data scientists based on whether they are able to use each explanation generated by the counterfactual versus Grad-CAM to correctly identify models which use SCs.
If the methods are successful on the human studies, the team has demonstrated the connection between them and the TUC of detecting positively correlated objects. }

\section{Conclusion}

Towards a diagnostic vision for IML, we presented a taxonomy as a way to clarify and begin bridging the gap between methods and use cases.
Further, we discussed best practices for how the taxonomy can be used and refined over time by researchers and consumers to better establish what methods are useful for what use cases.
As the taxonomy is flushed out via more studies by consumer-researcher teams, our vision is that it will be increasingly useful for both parties individually (Figure \ref{fig:abstractedversion}, middle). 
We hope that our discussions promote better practices in discovering, testing, and applying new and existing IML methods moving forward. 

\section{Acknowledgements}

We would like to thank David Alvarez-Melis, Maruan Al-Shedivat, Kasun Amarasinghe, Wenbo Cui, Lisa Dunlap, Boyang Fu, Rayid Ghani, Hoda Heidari, Oscar Li, Zack Lipton, Adam Perer, Marco Ribeiro, Kit Rodolfa, Sriram Sankararaman, Mukund Sundararajan, and Chih-Kuan Yeh for their valuable feedback. This work was supported in part by DARPA FA875017C0141, the National Science Foundation grants IIS1705121, IIS1838017 and IIS2046613, an Amazon Web Services Award, a Carnegie Bosch Institute Research Award, a Facebook Faculty Research Award, funding from Booz Allen Hamilton, and a Block Center Grant. Any opinions, findings and conclusions or recommendations expressed in this material are those of the author(s) and do not necessarily reflect the views of DARPA, the National Science Foundation, or any other funding agency.

\bibliographystyle{icml2021}
\bibliography{example_paper}

\begin{thebibliography}{48}
\providecommand{\natexlab}[1]{#1}
\providecommand{\url}[1]{\texttt{#1}}
\expandafter\ifx\csname urlstyle\endcsname\relax
  \providecommand{\doi}[1]{doi: #1}\else
  \providecommand{\doi}{doi: \begingroup \urlstyle{rm}\Url}\fi

\bibitem[Adebayo et~al.(2018)Adebayo, Gilmer, Muelly, Goodfellow, Hardt, and
  Kim]{adebayo2018sanity}
Adebayo, J., Gilmer, J., Muelly, M., Goodfellow, I., Hardt, M., and Kim, B.
\newblock Sanity checks for saliency maps.
\newblock In \emph{Advances in Neural Information Processing Systems}, pp.\
  9505--9515, 2018.

\bibitem[Alvarez-Melis \& Jaakkola(2018)Alvarez-Melis and
  Jaakkola]{alvarez2018robustness}
Alvarez-Melis, D. and Jaakkola, T.~S.
\newblock On the robustness of interpretability methods.
\newblock \emph{arXiv preprint arXiv:1806.08049}, 2018.

\bibitem[Ancona et~al.(2018)Ancona, Ceolini, {\"O}ztireli, and
  Gross]{ancona2018towards}
Ancona, M., Ceolini, E., {\"O}ztireli, C., and Gross, M.
\newblock Towards better understanding of gradient-based attribution methods
  for deep neural networks.
\newblock In \emph{International Conference on Learning Representations}, 2018.

\bibitem[Arya et~al.(2019)Arya, Bellamy, Chen, Dhurandhar, Hind, Hoffman,
  Houde, Liao, Luss, Mojsilovi{\'c}, et~al.]{arya2019one}
Arya, V., Bellamy, R.~K., Chen, P.-Y., Dhurandhar, A., Hind, M., Hoffman,
  S.~C., Houde, S., Liao, Q.~V., Luss, R., Mojsilovi{\'c}, A., et~al.
\newblock One explanation does not fit all: A toolkit and taxonomy of ai
  explainability techniques.
\newblock \emph{arXiv preprint arXiv:1909.03012}, 2019.

\bibitem[Bach et~al.(2015)Bach, Binder, Montavon, Klauschen, M{\"u}ller, and
  Samek]{bach2015pixel}
Bach, S., Binder, A., Montavon, G., Klauschen, F., M{\"u}ller, K.-R., and
  Samek, W.
\newblock On pixel-wise explanations for non-linear classifier decisions by
  layer-wise relevance propagation.
\newblock \emph{PloS one}, 10\penalty0 (7):\penalty0 e0130140, 2015.

\bibitem[Bansal et~al.(2020)Bansal, Wu, Zhu, Fok, Nushi, Kamar, Ribeiro, and
  Weld]{bansal2020does}
Bansal, G., Wu, T., Zhu, J., Fok, R., Nushi, B., Kamar, E., Ribeiro, M.~T., and
  Weld, D.~S.
\newblock Does the whole exceed its parts? the effect of ai explanations on
  complementary team performance.
\newblock \emph{arXiv preprint arXiv:2006.14779}, 2020.

\bibitem[Barocas et~al.(2020)Barocas, Selbst, and Raghavan]{barocas2020hidden}
Barocas, S., Selbst, A.~D., and Raghavan, M.
\newblock The hidden assumptions behind counterfactual explanations and
  principal reasons.
\newblock In \emph{Proceedings of the 2020 Conference on Fairness,
  Accountability, and Transparency}, pp.\  80--89, 2020.

\bibitem[Bhatt et~al.(2020)Bhatt, Xiang, Sharma, Weller, Taly, Jia, Ghosh,
  Puri, Moura, and Eckersley]{bhattpaper}
Bhatt, U., Xiang, A., Sharma, S., Weller, A., Taly, A., Jia, Y., Ghosh, J.,
  Puri, R., Moura, J. M.~F., and Eckersley, P.
\newblock Explainable machine learning in deployment.
\newblock In \emph{Proceedings of the 2020 Conference on Fairness,
  Accountability, and Transparency}, FAT* '20, pp.\  648–657, New York, NY,
  USA, 2020. Association for Computing Machinery.
\newblock ISBN 9781450369367.
\newblock \doi{10.1145/3351095.3375624}.
\newblock URL \url{https://doi.org/10.1145/3351095.3375624}.

\bibitem[Breiman(2001)]{breiman2001random}
Breiman, L.
\newblock Random forests.
\newblock \emph{Machine learning}, 45\penalty0 (1):\penalty0 5--32, 2001.

\bibitem[Chandrashekar \& Sahin(2014)Chandrashekar and
  Sahin]{chandrashekar2014survey}
Chandrashekar, G. and Sahin, F.
\newblock A survey on feature selection methods.
\newblock \emph{Computers \& Electrical Engineering}, 40\penalty0 (1):\penalty0
  16--28, 2014.

\bibitem[Chen et~al.(2018)Chen, Song, Wainwright, and Jordan]{chen2018learning}
Chen, J., Song, L., Wainwright, M., and Jordan, M.
\newblock Learning to explain: An information-theoretic perspective on model
  interpretation.
\newblock In \emph{International Conference on Machine Learning}, pp.\
  883--892, 2018.

\bibitem[Cook(1977)]{cook1977detection}
Cook, R.~D.
\newblock Detection of influential observation in linear regression.
\newblock \emph{Technometrics}, 19\penalty0 (1):\penalty0 15--18, 1977.

\bibitem[Doshi-Velez \& Kim(2017)Doshi-Velez and Kim]{doshivelez2017rigorous}
Doshi-Velez, F. and Kim, B.
\newblock Towards a rigorous science of interpretable machine learning, 2017.
\newblock URL \url{https://arxiv.org/abs/1702.08608}.

\bibitem[Friedman(2001)]{friedman2001greedy}
Friedman, J.~H.
\newblock Greedy function approximation: a gradient boosting machine.
\newblock \emph{Annals of statistics}, pp.\  1189--1232, 2001.

\bibitem[Frosst \& Hinton(2017)Frosst and Hinton]{frosst2017distilling}
Frosst, N. and Hinton, G.
\newblock Distilling a neural network into a soft decision tree.
\newblock \emph{arXiv preprint arXiv:1711.09784}, 2017.

\bibitem[Gilpin et~al.(2018)Gilpin, Bau, Yuan, Bajwa, Specter, and
  Kagal]{gilpin2018explaining}
Gilpin, L.~H., Bau, D., Yuan, B.~Z., Bajwa, A., Specter, M., and Kagal, L.
\newblock Explaining explanations: An overview of interpretability of machine
  learning.
\newblock In \emph{2018 IEEE 5th International Conference on data science and
  advanced analytics (DSAA)}, pp.\  80--89. IEEE, 2018.

\bibitem[Guidotti et~al.(2018)Guidotti, Monreale, Ruggieri, Turini, Giannotti,
  and Pedreschi]{guidotti2018survey}
Guidotti, R., Monreale, A., Ruggieri, S., Turini, F., Giannotti, F., and
  Pedreschi, D.
\newblock A survey of methods for explaining black box models.
\newblock \emph{ACM computing surveys (CSUR)}, 51\penalty0 (5):\penalty0 1--42,
  2018.

\bibitem[Hastie \& Tibshirani(1990)Hastie and
  Tibshirani]{hastie1990generalized}
Hastie, T.~J. and Tibshirani, R.~J.
\newblock \emph{Generalized additive models}, volume~43.
\newblock CRC press, 1990.

\bibitem[Hong et~al.(2020)Hong, Hullman, and Bertini]{hong2020human}
Hong, S.~R., Hullman, J., and Bertini, E.
\newblock Human factors in model interpretability: Industry practices,
  challenges, and needs.
\newblock \emph{Proceedings of the ACM on Human-Computer Interaction},
  4\penalty0 (CSCW1):\penalty0 1--26, 2020.

\bibitem[Hooker et~al.(2019)Hooker, Erhan, Kindermans, and
  Kim]{NEURIPS2019_fe4b8556}
Hooker, S., Erhan, D., Kindermans, P.-J., and Kim, B.
\newblock A benchmark for interpretability methods in deep neural networks.
\newblock In Wallach, H., Larochelle, H., Beygelzimer, A., d\textquotesingle
  Alch\'{e}-Buc, F., Fox, E., and Garnett, R. (eds.), \emph{Advances in Neural
  Information Processing Systems}, volume~32. Curran Associates, Inc., 2019.
\newblock URL
  \url{https://proceedings.neurips.cc/paper/2019/file/fe4b8556000d0f0cae99daa5c5c5a410-Paper.pdf}.

\bibitem[Kaur et~al.(2020)Kaur, Nori, Jenkins, Caruana, Wallach, and
  Wortman~Vaughan]{kaur2020interpreting}
Kaur, H., Nori, H., Jenkins, S., Caruana, R., Wallach, H., and Wortman~Vaughan,
  J.
\newblock Interpreting interpretability: Understanding data scientists' use of
  interpretability tools for machine learning.
\newblock In \emph{Proceedings of the 2020 CHI Conference on Human Factors in
  Computing Systems}, pp.\  1--14, 2020.

\bibitem[Kim et~al.(2016)Kim, Koyejo, Khanna, et~al.]{kim2016examples}
Kim, B., Koyejo, O., Khanna, R., et~al.
\newblock Examples are not enough, learn to criticize! criticism for
  interpretability.
\newblock In \emph{NIPS}, pp.\  2280--2288, 2016.

\bibitem[Koh \& Liang(2017)Koh and Liang]{koh2017understanding}
Koh, P.~W. and Liang, P.
\newblock Understanding black-box predictions via influence functions.
\newblock In \emph{International Conference on Machine Learning}, pp.\
  1885--1894, 2017.

\bibitem[Krishnan(2019)]{krishnan2019against}
Krishnan, M.
\newblock Against interpretability: a critical examination of the
  interpretability problem in machine learning.
\newblock \emph{Philosophy \& Technology}, pp.\  1--16, 2019.

\bibitem[Lakkaraju et~al.(2016)Lakkaraju, Bach, and
  Leskovec]{lakkaraju2016interpretable}
Lakkaraju, H., Bach, S.~H., and Leskovec, J.
\newblock Interpretable decision sets: A joint framework for description and
  prediction.
\newblock In \emph{Proceedings of the 22nd ACM SIGKDD international conference
  on knowledge discovery and data mining}, pp.\  1675--1684, 2016.

\bibitem[Laugel et~al.(2019)Laugel, Lesot, Marsala, and
  Detyniecki]{laugel2019issues}
Laugel, T., Lesot, M.-J., Marsala, C., and Detyniecki, M.
\newblock Issues with post-hoc counterfactual explanations: a discussion.
\newblock \emph{arXiv preprint arXiv:1906.04774}, 2019.

\bibitem[Li et~al.(2021)Li, Nagarajan, Plumb, and Talwalkar]{li2021a}
Li, J., Nagarajan, V., Plumb, G., and Talwalkar, A.
\newblock A learning theoretic perspective on local explainability.
\newblock In \emph{International Conference on Learning Representations}, 2021.
\newblock URL \url{https://openreview.net/forum?id=7aL-OtQrBWD}.

\bibitem[Lipton(2018)]{lipton2018mythos}
Lipton, Z.~C.
\newblock The mythos of model interpretability.
\newblock \emph{Queue}, 16\penalty0 (3):\penalty0 31--57, 2018.

\bibitem[Lou et~al.(2013)Lou, Caruana, Gehrke, and Hooker]{lou2013accurate}
Lou, Y., Caruana, R., Gehrke, J., and Hooker, G.
\newblock Accurate intelligible models with pairwise interactions.
\newblock In \emph{Proceedings of the 19th ACM SIGKDD international conference
  on Knowledge discovery and data mining}, pp.\  623--631, 2013.

\bibitem[Lundberg \& Lee(2017)Lundberg and Lee]{lundberg2017unified}
Lundberg, S.~M. and Lee, S.-I.
\newblock A unified approach to interpreting model predictions.
\newblock In \emph{Advances in neural information processing systems}, pp.\
  4765--4774, 2017.

\bibitem[Mohseni et~al.(2019)Mohseni, Zarei, and
  Ragan]{mohseni2019multidisciplinary}
Mohseni, S., Zarei, N., and Ragan, E.
\newblock A multidisciplinary survey and framework for design and evaluation of
  explainable ai systems. arxiv.
\newblock \emph{Human-Computer Interaction}, 2019.

\bibitem[Murdoch et~al.(2019)Murdoch, Singh, Kumbier, Abbasi-Asl, and
  Yu]{murdoch2019interpretable}
Murdoch, W.~J., Singh, C., Kumbier, K., Abbasi-Asl, R., and Yu, B.
\newblock Interpretable machine learning: definitions, methods, and
  applications.
\newblock \emph{arXiv preprint arXiv:1901.04592}, 2019.

\bibitem[Pawelczyk et~al.(2020)Pawelczyk, Broelemann, and
  Kasneci]{Pawelczyk_2020}
Pawelczyk, M., Broelemann, K., and Kasneci, G.
\newblock Learning model-agnostic counterfactual explanations for tabular data.
\newblock \emph{Proceedings of The Web Conference 2020}, Apr 2020.
\newblock \doi{10.1145/3366423.3380087}.
\newblock URL \url{http://dx.doi.org/10.1145/3366423.3380087}.

\bibitem[Plumb et~al.(2018)Plumb, Molitor, and Talwalkar]{plumb2018model}
Plumb, G., Molitor, D., and Talwalkar, A.~S.
\newblock Model agnostic supervised local explanations.
\newblock In \emph{Advances in Neural Information Processing Systems}, pp.\
  2515--2524, 2018.

\bibitem[Plumb et~al.(2020)Plumb, Terhorst, Sankararaman, and
  Talwalkar]{pmlr-v119-plumb20a}
Plumb, G., Terhorst, J., Sankararaman, S., and Talwalkar, A.
\newblock Explaining groups of points in low-dimensional representations.
\newblock In III, H.~D. and Singh, A. (eds.), \emph{Proceedings of the 37th
  International Conference on Machine Learning}, volume 119 of
  \emph{Proceedings of Machine Learning Research}, pp.\  7762--7771. PMLR,
  13--18 Jul 2020.
\newblock URL \url{http://proceedings.mlr.press/v119/plumb20a.html}.

\bibitem[Poyiadzi et~al.(2020)Poyiadzi, Sokol, Santos-Rodriguez, De~Bie, and
  Flach]{poyiadzi2020face}
Poyiadzi, R., Sokol, K., Santos-Rodriguez, R., De~Bie, T., and Flach, P.
\newblock Face: feasible and actionable counterfactual explanations.
\newblock In \emph{Proceedings of the AAAI/ACM Conference on AI, Ethics, and
  Society}, pp.\  344--350, 2020.

\bibitem[Rawal \& Lakkaraju(2020)Rawal and Lakkaraju]{rawal2020beyond}
Rawal, K. and Lakkaraju, H.
\newblock Beyond individualized recourse: Interpretable and interactive
  summaries of actionable recourses.
\newblock \emph{Advances in Neural Information Processing Systems}, 33, 2020.

\bibitem[Ribeiro et~al.(2016)Ribeiro, Singh, and Guestrin]{ribeiro2016should}
Ribeiro, M.~T., Singh, S., and Guestrin, C.
\newblock " why should i trust you?" explaining the predictions of any
  classifier.
\newblock In \emph{Proceedings of the 22nd ACM SIGKDD international conference
  on knowledge discovery and data mining}, pp.\  1135--1144, 2016.

\bibitem[Ribeiro et~al.(2018)Ribeiro, Singh, and Guestrin]{ribeiro2018anchors}
Ribeiro, M.~T., Singh, S., and Guestrin, C.
\newblock Anchors: High-precision model-agnostic explanations.
\newblock In \emph{Proceedings of the AAAI conference on artificial
  intelligence}, volume~32, 2018.

\bibitem[Rudin(2019)]{rudin2019stop}
Rudin, C.
\newblock Stop explaining black box machine learning models for high stakes
  decisions and use interpretable models instead.
\newblock \emph{Nature Machine Intelligence}, 1\penalty0 (5):\penalty0
  206--215, 2019.

\bibitem[Selvaraju et~al.(2017)Selvaraju, Cogswell, Das, Vedantam, Parikh, and
  Batra]{selvaraju2017grad}
Selvaraju, R.~R., Cogswell, M., Das, A., Vedantam, R., Parikh, D., and Batra,
  D.
\newblock Grad-cam: Visual explanations from deep networks via gradient-based
  localization.
\newblock In \emph{Proceedings of the IEEE international conference on computer
  vision}, pp.\  618--626, 2017.

\bibitem[Simonyan et~al.(2013)Simonyan, Vedaldi, and
  Zisserman]{simonyan2013deep}
Simonyan, K., Vedaldi, A., and Zisserman, A.
\newblock Deep inside convolutional networks: Visualising image classification
  models and saliency maps.
\newblock \emph{arXiv preprint arXiv:1312.6034}, 2013.

\bibitem[Sundararajan et~al.(2017)Sundararajan, Taly, and
  Yan]{sundararajan2017axiomatic}
Sundararajan, M., Taly, A., and Yan, Q.
\newblock Axiomatic attribution for deep networks.
\newblock In \emph{International Conference on Machine Learning}, pp.\
  3319--3328, 2017.

\bibitem[Ustun et~al.(2019)Ustun, Spangher, and Liu]{Ustun_2019}
Ustun, B., Spangher, A., and Liu, Y.
\newblock Actionable recourse in linear classification.
\newblock \emph{Proceedings of the Conference on Fairness, Accountability, and
  Transparency}, Jan 2019.
\newblock \doi{10.1145/3287560.3287566}.
\newblock URL \url{http://dx.doi.org/10.1145/3287560.3287566}.

\bibitem[Wang \& Rudin(2015)Wang and Rudin]{wang2015falling}
Wang, F. and Rudin, C.
\newblock Falling rule lists.
\newblock In \emph{Artificial Intelligence and Statistics}, pp.\  1013--1022,
  2015.

\bibitem[Williamson \& Feng(2020)Williamson and Feng]{pmlr-v119-williamson20a}
Williamson, B. and Feng, J.
\newblock Efficient nonparametric statistical inference on population feature
  importance using shapley values.
\newblock In \emph{Proceedings of the 37th International Conference on Machine
  Learning}, pp.\  10282--10291, 2020.

\bibitem[Yeh et~al.(2018)Yeh, Kim, Yen, and Ravikumar]{yeh2018representer}
Yeh, C.-K., Kim, J., Yen, I. E.-H., and Ravikumar, P.~K.
\newblock Representer point selection for explaining deep neural networks.
\newblock In \emph{Advances in neural information processing systems}, pp.\
  9291--9301, 2018.

\bibitem[Zhang et~al.(2018)Zhang, Solar-Lezama, and Singh]{zhangnips}
Zhang, X., Solar-Lezama, A., and Singh, R.
\newblock Interpreting neural network judgments via minimal, stable, and
  symbolic corrections.
\newblock In Bengio, S., Wallach, H., Larochelle, H., Grauman, K.,
  Cesa-Bianchi, N., and Garnett, R. (eds.), \emph{Advances in Neural
  Information Processing Systems 31}, pp.\  4874--4885. Curran Associates,
  Inc., 2018.

\end{thebibliography}

\clearpage
\appendix{}

\section{Taxonomy of Method Goals}
\label{appendix:taxonomy}

\begin{enumerate}
    \item \emph{Explanation representation}. 
    Model explanations are typically given in terms of either \textit{feature relationships} between inputs and outputs or \textit{training examples}. 
    \item \emph{Type of feature relationships}. 
    In the context of explanations based on feature relationships, there are three distinct approaches for explaining different aspects of the model's reasoning: \textit{feature attribution}, \textit{counterfactual}, and \textit{approximation}. As a note, due to there being less focus from the IML community on training example-based explanations, we consider one main grouping along that branch,  \textit{sample importance} explanations.
    \item \emph{Explanation scale}. Explanations vary in terms of the scale of the desired insights, with their scope ranging from  how \textit{local} (i.e. for an individual instance) to \textit{global} (i.e. for a well defined region of the input space). 
\end{enumerate}

\section{Method Cluster Details}
\label{appendix:methodclusters}

\textit{\underline{Feature attribution}} methods address the question of how features present (or missing) in the input(s) affect the model's prediction(s) (i.e. how ``important'' each feature is to the model's prediction(s)). Often,  measures of importance are defined based on how the model's prediction(s) change relative to its prediction for some baseline input. The baseline input is sometimes implicit and is typically domain specific (e.g. all black pixels for grayscale images or the mean input in tabular data).
Thus, the technical details here are both the precise \textit{notion of “importance”} as well as the choice of the \textit{baseline input}.

\textit{Local feature attribution} methods like SHAP \cite{lundberg2017unified} attribute the change in the conditional expectation of the model output conditioned on the features of interest, with respect to an explicit baseline input. 
Therefore the output explanation differs significantly based on the baseline input chosen. 
Other methods such as Grad-CAM \cite{selvaraju2017grad} and Integrated Gradients \cite{sundararajan2017axiomatic}, treat gradients and their variants as the importance values, and thus are usually restricted to deep neural networks. The latter, like SHAP, also requires carefully choosing an explicit baseline input. 
Separately, a family of methods, such as L2X \cite{chen2018learning}, use mutual information between the features and labels to learn the importance values. 
Relevant proxy metrics typically measure how much the model prediction changes for different types of perturbations applied to the individual (or the training data) according to the ``importance'' values as computed by each method~\cite{bach2015pixel, alvarez2018robustness, ancona2018towards, NEURIPS2019_fe4b8556}.

\textit{Global feature attribution} methods include traditional feature selection approaches from classical statistics \cite{chandrashekar2014survey} or model specific approaches tailored to specific classes such as decision trees and tree ensembles \cite{friedman2001greedy, breiman2001random}. 
Because these approaches are either computationally expensive or model-specific, more recent methods focus on aggregating local feature attributions, such as how \cite{pmlr-v119-williamson20a} estimates the Shapley-based metric, SPVIM. 
Proxy metrics might integrate over the domain at the individual-level to derive a global-level measure.

\underline{\textit{Counterfactual}} explanations identify a “low cost” modification that can be applied to data point(s) to get a different prediction. 
The most common technical detail is the specific measure of \textit{cost} and the most common proxy metric is how often the counterfactual changes the model's prediction(s). 

\textit{Local Counterfactual} methods includes POLARIS \cite{zhangnips}, which finds stable counterfactual points (i.e. where the larger region around it also has a different prediction). 
Meanwhile, FACE \cite{poyiadzi2020face} tries to find a counterfactual that is on the data distribution and is thus realistic to change into, which might be an important requirement for real-world applications (i.e. the feature that represents a person's ``income'' cannot be easily doubled for a potentially better mortgage rate if that person does not have the capacity to do so). As such, another set of proxy metrics typically tries to capture how real or feasible the proposed changes are with the amount of cost incurred for individual instances, as well as the distribution of these costs over different sub-groups of the data~\cite{Ustun_2019, Pawelczyk_2020}. 


\textit{Global counterfactual} explanations finds a modification that can be applied to a whole group of points. 
For example, ELDR \cite{pmlr-v119-plumb20a} identifies which features (genes in its original medical use case) differentiate different clusters of data (cell types), and AReS \cite{rawal2020beyond} aims to do this to detect model bias. 
One important proxy metric to consider for global methods is coverage~\cite{pmlr-v119-plumb20a}, which measures the degree to which the explanations capture all of the differences between different cluster of points.  

\underline{\textit{Approximation}} methods aim to use a simple function to approximate the model’s behavior as accurately as possible in a region, either locally around a data point or globally around as many points as possible or across a specific region of the input space. 
These methods require the technical detail of both what that \textit{region} is and what the simple function’s \textit{model family} is.

\textit{Local approximation} methods are most well known by its canonical method, LIME \cite{ribeiro2016should}, which weights data points drawn uniformly from an interpretable feature representation using their similarity to the point being explained. 
Other methods such as MAPLE \cite{plumb2018model} leverage the structure of the underlying data distribution to generate local approximations. 
One canonical proxy metric is local fidelity~\cite{plumb2018model, li2021a}, which measures how well the approximation method predicts within a certain neighborhood of data points. 

\textit{Global approximation} methods include distillation \cite{frosst2017distilling}, which leverage the more intuitive representation of models such as shallow decision trees to approximate a more complex model’s decision process. 
Another method, Generalized additive model (GAM), and its variant (GA2M) \cite{lou2013accurate} benefit from being able to represent a prediction in terms of univariate features and pairwise interactions. 
Finally, a third model type includes falling rule lists \cite{wang2015falling}, decision sets \cite{lakkaraju2016interpretable}, and Anchors \cite{ribeiro2018anchors}, which create lists of if-then rules on the features that best replicate the model’s decision process. 
A canonical proxy metric is coverage~\cite{ribeiro2018anchors}, which in this context measures how many data points the explanation applies to.

\underline{\textit{Sample importance}} methods aim to understand how either model's prediction on an individual point or the model as a whole is impacted by changes in the training data. 
Technical details differ from method to method, so currently it is difficult to identify a uniform axis of variation. 
These methods can be evaluated with proxy metrics that represent the usefulness of the provided explanations, through simulated experiments of finding corrupted data points~\cite{yeh2018representer}, detecting points responsible data distribution shifts~\cite{koh2017understanding}, and recovering high accuracy with the samples considered important~\cite{kim2016examples}.

\textit{Local sample importance} methods include influence functions \cite{cook1977detection, koh2017understanding}, which compute the effect of removing or perturbing a training point on the resulting model’s loss for a particular test point.
Meanwhile, representer point selection \cite{yeh2018representer} decomposes the model prediction value on the test point in terms of the neural network activations of each training sample, computing a similar notion of influence but in a different manner. 
Such methods have been shown to be effective for dataset debugging~\cite{yeh2018representer} and detecting vulnerable examples for dataset poisoning~\cite{koh2017understanding}.

\textit{Global sample importance} methods, on the other hand, compute the effect of removing or perturbing a training point on the model's learned parameters and does not require the specification of a test point. Influence functions~\cite{koh2017understanding} do this by approximating the Hessian of the loss for the training point, while representer point selection~\cite{yeh2018representer} explicitly decomposes the weights as the linear combination of the training point activations. 

\newpage
\subsection{More on Cross-cluster Adaptation}
\label{appendix:crosscluster}

\begin{table}[h!]
\caption{
We discuss important limitations and assumptions one should consider when performing cross-cluster adaptations of one method cluster to answer a TO of another cluster, specifically for local explanations that are feature attributions (FA), counterfactuals (CF), and approximations (AP).}
\begin{tabular}{|l|l|}
\hline
\textbf{FA → AP}  & \begin{tabular}[c]{@{}l@{}}It is unclear how one should map FA scores \\ to the ``parameters'' in an approximation. \\ For instance,  one might attempt to use \\ importance scores as linear coefficients, \\ but this will not work in general.\end{tabular}                                                                                                                                          \\ \hline
\textbf{FA → CF}      & \begin{tabular}[c]{@{}l@{}}One could possibly adapt FA methods to \\ do CFs, for example, by changing the most \\  important features to their baseline input.  \\ However, it is likely that the resulting point \\ is not very close to the original or not very \\ realistic and, as a result, may do poorly on \\ the `‘low cost’' part of the CF objective.\end{tabular} \\ \hline
\textbf{AP → CF}  & \begin{tabular}[c]{@{}l@{}}One could adapt APs by computing a CF \\ on  the surrogate model, which might \\ be easier than on the complex full model. \\ However, there is no guarantee  these  CFs \\ hold exactly on the original model given \\ the surrogate model is an approximation.\end{tabular}                                                          \\ \hline
\textbf{AP →  FA} & \begin{tabular}[c]{@{}l@{}}One could adapt APs to derive FA scores  \\ by simply using weights derived from a \\ surrogate model, say the coefficients of a  \\ linear approximation. Its success would \\ depend on how close the intended baseline \\ input of the FA is to the neighborhood \\ region used  by the approximation.\end{tabular}                                                                                                                                                                          \\ \hline
\textbf{CF → FA}      & \begin{tabular}[c]{@{}l@{}} One could use the CF  perturbation in \\ feature space and derive FA scores by \\ saying the features that are  changed  are the \\ most important. However, this  also depends \\ on matching the intended baseline  input(s)\\ and the point(s) one generates the CF for. \end{tabular}                                                                                                                                                           \\ \hline
\textbf{CF → AP}   & \begin{tabular}[c]{@{}l@{}}A single CF for a single original point \\ is likely insufficient to approximate the \\ function for non-trivial data dimensions. \\ However, it may be possible for one to use  \\ a diverse set of CFs for the same point.\end{tabular}                                                                                                              \\ \hline
\end{tabular}
\label{table:adapation}
\end{table}

\end{document}